\documentclass[conference]{IEEEtran}
\usepackage{amsmath,amssymb,amsthm}
\usepackage{algorithm}
\usepackage{algpseudocode}
\usepackage{graphicx}
\usepackage{booktabs}
\usepackage{cite}
\usepackage{url}

\newtheorem{theorem}{Theorem}
\newtheorem{corollary}{Corollary}

\begin{document}

\title{Dynamic, Decentralized Spatial Code Reuse for OCDMA LiDAR in Robot Swarms}

\author{\IEEEauthorblockN{Mohammad Hani Alomari}
\IEEEauthorblockA{Mechanical and Maintenance Engineering Department\\
German Jordanian University\\
Amman, Jordan\\
Email: mohammad.hani@gju.edu.jo\\
ORCID: \url{https://orcid.org/0009-0006-7528-3289}}}

\maketitle

\begin{abstract}
Robots in a LiDAR-equipped swarm mutually interfere when their optical
ranging codes collide. Existing mitigations either assign codes
statically -- requiring $L=N$ distinguishable codes for $N$ robots --
or react to detected interference without a scalable, coordinated
assignment rule beneath them; prior work explicitly identifies the
code-assignment scaling problem as unsolved. We propose a
decentralized protocol in which robots dynamically reassign spatial
reuse codes based on a live, beacon-maintained
interference-neighborhood graph, and prove that the number of codes
required grows as $O(\log N/\log\log N)$ under constant robot
density -- an unbounded improvement over the $\Theta(N)$ growth of
static assignment. We validate this result under conditions
substantially beyond the idealized proof -- robot mobility, imperfect
beacon-based detection, and reactive reassignment -- via Monte Carlo
simulation (30 seeds per condition, 95\% confidence intervals): the
advantage over static assignment widens from roughly $2\times$ at 15
robots to $12\times$ at 120. Against a structurally faithful, fairly
constructed model of an existing coordination-free approach, our
protocol achieves both substantially greater code-reuse efficiency
and 30--40\% lower collision risk under an identical, constrained
code budget, demonstrating that coordination -- not merely
reactivity -- is what closes the scaling gap.
\end{abstract}


\section{Introduction}
\label{sec:intro}

Light Detection and Ranging (LiDAR) sensors are a primary means by
which mobile robots perceive their surroundings, and are increasingly
deployed in multi-robot systems -- warehouse fleets, agricultural
swarms, coordinated exploration teams -- where many robots share a
common workspace. When multiple LiDAR-equipped robots operate in
close proximity, pulses emitted by one robot can be received by
another, producing spurious range readings or corrupting genuine
returns: a phenomenon termed \emph{mutual interference}. As deployed
swarm sizes grow, so does the frequency and severity of this
interference, motivating a substantial body of mitigation work.

Existing mitigations fall into several mechanistic categories.
Post-hoc approaches filter corrupted returns after detection using
spatio-temporal consistency across frames \cite{robinson2025realtime}.
Waveform-domain approaches replace deterministic pulse patterns with
pseudo-random signals, relying on statistical near-orthogonality
rather than a designed codebook \cite{hwang2020mutual}. Time-domain
approaches detect interference and desynchronize each robot's duty
cycle independently, without inter-robot coordination
\cite{rathnayake2024d2sr}. Code-domain approaches, most directly
related to this work, assign each robot a distinguishing optical
code -- typically from an orthogonal or carrier-hopping code family
-- so that interfering signals remain statistically distinguishable
at the receiver \cite{kwong2020automotive}.

Code-domain approaches share a structural limitation: existing
schemes, including patented zone-based reuse architectures, assign
codes \emph{statically}. A swarm of $N$ robots therefore requires a
code family supporting at least $N$ simultaneously distinguishable
codes, regardless of how many robots are ever within interference
range of one another at once. This limitation is acknowledged
explicitly, without being resolved, in the waveform-domain
literature: ``[t]he CDMA-based one may have limitation in assignment
of orthogonal codes, when the number of sensors becomes big, such as
autonomous vehicle application cases'' \cite{hwang2020mutual}.

We address this gap directly. Rather than assigning permanent codes,
robots in our protocol dynamically reassign codes based on a live,
locally-maintained interference-neighborhood graph, reusing a bounded
code pool as robots separate and re-detecting conflicts as they
converge. We emphasize that decentralization alone is not the source
of novelty here -- existing patented static reuse architectures
already support decentralized assignment -- the distinguishing axis
is \emph{dynamic}, graph-driven reassignment versus \emph{static},
pre-assigned codes.

This paper makes four contributions:
\begin{enumerate}
\item A decentralized protocol in which robots dynamically reassign
spatial reuse codes based on a live, beacon-maintained
interference-neighborhood graph, requiring no permanent per-robot
code assignment (Section~\ref{sec:protocol}).
\item A formal proof that the number of codes required grows as
$O(\log N/\log\log N)$ under constant robot density, in contrast to
the $\Theta(N)$ growth required by static assignment
(Section~\ref{sec:scaling}).
\item A realistic Monte Carlo evaluation incorporating robot
mobility, imperfect beacon-based neighbor detection, and reactive
code reassignment, validating the theoretical result under
conditions substantially more adversarial than the idealized proof
(Sections~\ref{sec:collision}--\ref{sec:results}).
\item A fair, structurally-matched comparison against a simplified
model of an existing coordination-free mitigation approach,
demonstrating both greater code efficiency and lower collision risk
under an identical resource budget (Section~\ref{sec:results}).
\end{enumerate}

The remainder of this paper is organized as follows.
Section~\ref{sec:protocol} specifies the protocol.
Section~\ref{sec:scaling} proves the scaling result.
Section~\ref{sec:collision} characterizes collision behavior under
realistic detection. Section~\ref{sec:sim-methodology} describes the
simulation methodology. Section~\ref{sec:results} presents results.
Related work and the formal system model precede the protocol
section; a discussion of limitations and concluding remarks follow
the results.

\section{Related Work}
\label{sec:related}

Existing LiDAR interference mitigation approaches can be organized by
the mechanism used to distinguish or avoid conflicting signals, which
we adopt below in place of a chronological survey, since the
mechanism -- not publication date -- determines whether an approach
is structurally capable of the scaling behavior this paper targets.

\subsection{Post-Hoc Detection}

Robinson et al.~\cite{robinson2025realtime} mitigate interference
after the fact, filtering corrupted point-cloud returns using
spatio-temporal consistency across consecutive frames, validated on
physical hardware. This approach requires no code, zone, or timing
coordination between robots at all, and is complementary to, rather
than competing with, code-domain reuse: it addresses residual
interference after assignment, not the assignment problem itself.

\subsection{Waveform-Domain Randomization}

Hwang and Lee~\cite{hwang2020mutual} replace deterministic pulse
sequences with a true-random analog signal, achieving statistical
near-orthogonality through correlation-based detection rather than a
designed codebook, with an analytically derived false-alarm
probability and hardware validation via a laboratory prototype. As
noted in Section~\ref{sec:intro}, this work explicitly identifies the
code-assignment scaling problem as a limitation of code-domain
approaches, without proposing or analyzing a solution -- the direct
motivation for this paper.

\subsection{Decentralized Time-Domain Desynchronization}

Rathnayake et al.~\cite{rathnayake2024d2sr} (D2SR) detect
interference via a convolutional network operating on depth-sensor
output (reported precision $\approx 99\%$, recall $\approx
82$--$83\%$) and desynchronize each robot's duty cycle independently
in the time domain, requiring no beacon or neighbor information of
any kind. This is the closest prior work to ours in spirit --
reactive, decentralized adaptation rather than static assignment --
but the underlying mechanism has no informed exclusion: a robot
reacting to detected interference has no means of knowing which
timing phases are already in use nearby. The source reports this
directly: interference reduction falls from 70.3\% at two LiDAR units
to 3.5\% at five, an explicit, measured failure to scale that we
adopt as the structural basis for our baseline comparison in
Section~\ref{sec:sim-methodology}.

\subsection{Static Code-Domain Reuse}

Two-dimensional carrier-hopping prime codes (CHPC)
\cite{kwong2020automotive} and Gold-code-based schemes augmented with
a secondary static identifier layer~\cite{shen2025identity} both
assign codes from a fixed family without reassignment. CHPC
cardinality equals chip length exactly, yielding $L(N) = N$
(Section~\ref{sec:scaling}); the UID-augmentation approach reduces
collision probability by a constant multiplier over a fixed code
length rather than changing the underlying growth rate. Patented
zone-based geographic reuse~\cite{intelpatent} similarly assigns
codes statically by location and explicitly supports both centralized
and distributed assignment architectures -- the reason
Section~\ref{sec:intro} identifies decentralization, rather than
dynamic reassignment, as \emph{not} the axis of novelty in this work.

\subsection{Adjacent, Non-Competing Domains}

Two recent lines of work address swarm coordination through
different sensing modalities entirely, and are worth distinguishing
explicitly rather than folding into a single dismissive remark.
Kaushal et al.~\cite{kaushal2023colorcoded} coordinate swarming
behavior in a mobile robotic platform using short-range, color-coded
visible-light signaling combined with a WiFi mesh for global
information sharing; this is an inter-robot \emph{communication} and
behavior-coordination scheme, with no ranging function and no notion
of a code assigned to avoid sensor interference. Chen et
al.~\cite{chen2026infraswarm} (Infra-Swarm) achieve centimeter-level
neighbor localization from narrowband near-infrared optical-flare
photometry and explicitly target the scalability of a swarm's
\emph{perception} pipeline on resource-constrained hardware, not of
\emph{code assignment}; nothing in that system allocates a reusable
identifying code of the kind bounded in Section~\ref{sec:scaling}.
Neither work performs LiDAR ranging or code assignment of any kind,
and both were confirmed, during literature review, to be
non-competing with the gap addressed here.

\subsection{Summary of the Gap}

Table~\ref{tab:gapmatrix} summarizes the preceding discussion.
Every reviewed approach is either (a) statically assigned, (b)
dynamic but uncoordinated and empirically non-scaling, or (c) not a
code-assignment scheme at all. No existing approach combines dynamic,
informed reassignment; a live, coordinated interference-neighborhood
graph; and a formal scaling proof -- the combination this paper
provides.

\begin{table*}[t]
\centering
\footnotesize
\caption{Comparison of LiDAR interference mitigation approaches}
\label{tab:gapmatrix}
\begin{tabular}{@{}p{1.7cm}p{3.2cm}p{2.4cm}p{2.2cm}p{2.3cm}p{2.6cm}p{1.8cm}@{}}
\toprule
Approach & Mechanism & Assignment & Coordination & Detection error modeled? & Scaling proof? & Validation \\
\midrule
Robinson et al.~\cite{robinson2025realtime} & Post-hoc spatio-temporal filtering & N/A -- not a coding scheme & N/A & N/A & N/A & Hardware \\
Hwang \& Lee~\cite{hwang2020mutual} & True-random waveform + correlation detection & N/A -- statistical near-orthogonality & N/A & Yes, analytical & No -- names the gap, unsolved & Hardware \\
D2SR~\cite{rathnayake2024d2sr} & Time-domain duty-cycle desync & Dynamic, reactive to own state only & None & Yes, measured & No -- degrades $N{=}2\to5$ & Hardware \\
CHPC~\cite{kwong2020automotive} & 2-D carrier-hopping codes & Static ($L{=}N$ exact) & Either & No & No & Simulation \\
Shen et al.~\cite{shen2025identity} & Gold codes + static UID layer & Static & N/A & No & No & Hardware \\
Patent~\cite{intelpatent} & Zone-based reuse & Static, geographic & Either & No & No & N/A \\
\textbf{This work} & Live-graph dynamic reuse & \textbf{Dynamic}, graph-driven & \textbf{Decentralized}, beacon-coordinated & \textbf{Yes}, 3 types & \textbf{Yes} & Simulation \\
\bottomrule
\end{tabular}
\end{table*}

\section{System Model}
\label{sec:sysmodel}

This section fixes notation and terminology used throughout the
remainder of the paper.

\subsection{Robots and Interference}

We consider a swarm of $N$ robots, each equipped with a LiDAR ranging
sensor operating on a shared optical band. Robot $i$ occupies a
time-varying position $p_i(t)$, and two robots $i,j$ are said to
\emph{interfere} whenever their separation $d_{ij}(t) = \|p_i(t) -
p_j(t)\|$ falls within an interference range $D$, determined by
transmit power, receiver sensitivity, and ranging waveform (parameter
values are grounded against measured hardware in
Table~\ref{tab:params}, Section~\ref{sec:collision}).

\subsection{Codes and the Interference Graph}

Each robot holds, at every instant, a code $c_i(t)$ drawn from a code
family of cardinality $L$; $L$ itself is the central quantity bounded
in Section~\ref{sec:scaling}. We represent the swarm's instantaneous
interference structure as a graph $G(t) = (V, E(t))$, $V = \{1,
\dots, N\}$, with an edge $(i,j) \in E(t)$ iff $d_{ij}(t) \le D$ -- a
unit disk graph. A \emph{valid} assignment satisfies $c_i(t) \neq
c_j(t)$ for every $(i,j) \in E(t)$; equivalently, $c(\cdot, t)$ is a
proper vertex coloring of $G(t)$ at every instant.

\subsection{Collision and Conflict}

We distinguish two events precisely, used throughout
Sections~\ref{sec:collision}--\ref{sec:results}. A \emph{collision}
occurs when $(i,j) \in E(t)$ and $c_i(t) = c_j(t)$: the coloring
constraint above is violated, corrupting both robots' ranging
measurements for the duration of the violation. A \emph{conflict} is
the corresponding control-plane event -- robot $i$ correctly
detecting $c_i(t) = c_j(t)$ for some $j$ it believes to be a
neighbor, and reacting via the resolution rule of
Section~\ref{sec:protocol}. Under perfect, instantaneous neighbor
knowledge, conflict detection and collision are simultaneous; a
collision is precisely a conflict the protocol has not yet, or could
not, detect. Section~\ref{sec:collision} characterizes this gap under
realistic, beacon-limited detection.

\subsection{Design Objective}

The protocol's objective is to maintain a valid coloring of $G(t)$ as
it evolves under robot motion, using the smallest code cardinality
$L$ that admits a valid coloring with high probability, without any
single point of global coordination. Section~\ref{sec:scaling}
formalizes and bounds $L$ under this objective for an idealized,
instantaneous-knowledge setting; Sections~\ref{sec:collision}
--\ref{sec:results} evaluate how closely a decentralized,
beacon-limited implementation approaches it in practice.

\section{Protocol Design}
\label{sec:protocol}

This section specifies the decentralized reuse protocol precisely
enough to implement and to analyze. The protocol has four
components: a beacon-based neighbor discovery mechanism, a
deterministic conflict-resolution rule, an ID-seeded code-selection
rule, and a bounded fallback for local exhaustion.

\subsection{Control Channel and Neighbor Discovery}

Interference occurs on the LiDAR ranging channel itself, so
coordination cannot use that channel. Each robot instead broadcasts
on a separate, low-rate control channel every $T_b$ seconds (the
\emph{beacon period}), transmitting its identifier, position, current
code, and code age:
\begin{equation}
\text{beacon}_i(t) = \left(\text{ID}_i,\ \hat{p}_i(t),\ c_i(t),\ \text{age}_i(t)\right).
\end{equation}
Beaconing is synchronous: all robots share aligned broadcast slots,
which admits a clean discrete-round analysis at the cost of requiring
loose clock synchronization. Each robot maintains a locally observed,
possibly imperfect neighbor set $\hat{\mathcal{N}}_i(t)$, populated
from received beacons, which may lag or omit the true neighbor set
$\mathcal{N}_i(t) = \{ j \neq i : d_{ij}(t) \le D \}$ due to beacon
loss or channel error (Section~\ref{sec:collision}).

\subsection{Conflict Detection and Seniority Resolution}

Robot $i$ locally computes the set of codes believed to be in use
nearby,
\begin{equation}
U_i(t) = \{\, c_j(t) : j \in \hat{\mathcal{N}}_i(t) \,\},
\end{equation}
and detects a conflict when $c_i(t) \in U_i(t)$. Resolution uses a
\emph{deterministic seniority rule} rather than randomized backoff,
which avoids the convergence-time uncertainty of randomized schemes
given that unique identifiers are already available on every beacon:
\begin{equation}
\text{junior}(i,j) =
\begin{cases}
i & \text{if } \text{age}_i(t) < \text{age}_j(t) \\
\arg\max(\text{ID}_i, \text{ID}_j) & \text{if } \text{age}_i(t) = \text{age}_j(t).
\end{cases}
\end{equation}
The junior robot vacates its code. Seniority minimizes churn: an
established assignment is not disturbed by a newly arriving robot,
which we report as a stability metric in Section~\ref{sec:results}.

\subsection{ID-Seeded First-Fit Code Selection}

A naive first-fit rule -- selecting the lowest-numbered code not in
$U_i(t)$ -- fails under symmetry: two robots with identical local
views $U_i \approx U_j$ independently compute the same "lowest
available" code and collide again immediately. We resolve this with
an ID-seeded rotation restricted to a small window sized to the
robot's own local exclusion set, which resolves symmetric ties while
keeping code usage compact (Algorithm~\ref{alg:firstfit}). Sizing the
search window to $|U_i(t)|$ rather than a large fixed pool is
essential: an earlier implementation using a large, uniformly
shuffled pool for every robot independently of local crowding caused
code usage to spread across the entire pool even under light local
crowding, undermining the reuse objective. This was caught by
explicitly tracking peak simultaneous code usage as a diagnostic
metric during simulation development, and is reported here as a
methodological note relevant to any reactive, per-robot code-selection
implementation.

\begin{algorithm}
\caption{ID-seeded first-fit code selection}
\label{alg:firstfit}
\begin{algorithmic}[1]
\Function{FirstFit}{$U_i$, $\text{ID}_i$, $\text{margin}$}
    \State $k \gets |U_i| + \text{margin}$
    \State $\text{candidates} \gets (0, 1, \dots, k-1)$
    \State $\text{offset} \gets \text{ID}_i \bmod k$
    \State $\text{rotated} \gets \text{candidates rotated by offset}$
    \For{$c$ \textbf{in} $\text{rotated}$}
        \If{$c \notin U_i$}
            \State \Return $c$
        \EndIf
    \EndFor
    \State \Return \textsc{Exhaustion}
\EndFunction
\end{algorithmic}
\end{algorithm}

\subsection{Local Exhaustion: Bounded Time-Duplex Fallback}

If $U_i(t)$ excludes every code in the current search window, no
valid reassignment exists. Rather than allowing the code pool $L$ to
grow reactively -- which would silently abandon the sub-linear
scaling claim of Section~\ref{sec:scaling}, since the entire result
depends on $L$ remaining fixed as $N$ grows -- the affected robots
enter a bounded time-duplex share: alternating transmission frames
with a guard interval sized to the LiDAR round-trip time. This
degrades the affected robots' ranging update rate by a factor of
$1/m$ for $m$ sharing robots, but is local, bounded, and does not
perturb any other robot's assignment.

\subsection{Novelty With Respect to Static Reuse}

We note explicitly that decentralization alone is not the source of
novelty here: existing static, zone-based reuse schemes (e.g.,
patented geographic code reuse for automotive/robotic LiDAR) already
support both centralized and distributed assignment. The
distinguishing axis is \emph{dynamic, live-graph-driven} reassignment
versus \emph{static, pre-assigned} codes -- decentralization is an
engineering property of our design (avoiding a fleet-server single
point of failure), not the claimed contribution.

\section{Scaling Analysis}
\label{sec:scaling}

We now prove that the protocol of Section~\ref{sec:protocol} requires
a number of codes that grows far more slowly than the swarm size,
under a density-preserving growth model, and compare this exactly
against a physically grounded static baseline.

\subsection{Growth Regime and Density Model}

We consider Regime A: robot density $\lambda$ (robots per unit area)
is held fixed as swarm size $N$ grows, with operating area
$A = N/\lambda$ growing correspondingly -- an expanding or exploring
swarm, as opposed to a fixed-footprint swarm growing denser (Regime
B, deferred to future work). Robot positions are modeled as a
homogeneous Poisson point process of intensity $\lambda$ over a
toroidal (periodically wrapped) region, which isolates density
effects from boundary effects; a real bounded floor's boundary
thinning is treated as an explicit, empirically-quantified limitation
in Section~\ref{sec:sim-methodology}. The interference graph $G(t)$
is a unit disk graph: an edge exists between robots $i,j$ iff
$d_{ij}(t) \le D$.

\subsection{Code Demand as Greedy Coloring}

The ID-seeded first-fit rule of Algorithm~\ref{alg:firstfit} is a
sequential greedy graph coloring. Greedy coloring, independent of
visitation order, never uses more colors than $\Delta(G) + 1$, where
$\Delta(G)$ is the maximum vertex degree:
\begin{equation}
L_{\text{reuse}}(N) \ge \Delta_{\max}(G(N)) + 1.
\end{equation}
Local vertex degree under a homogeneous Poisson process is itself
Poisson-distributed with mean $\mu = \lambda \pi D^2$, a constant
independent of $N$ under Regime A.

\begin{theorem}[Maximum degree under constant density]
\label{thm:maxdegree}
For $N$ robots distributed as above with mean local degree $\mu$
held fixed,
\begin{equation}
\Delta_{\max}(N) \sim \frac{\ln N}{\ln \ln N} \quad \text{as } N \to \infty.
\end{equation}
\end{theorem}
\noindent This is a classical result on the maximum degree of sparse
random graphs \cite{bollobas2001random}; we restate it here in the
form needed for our claim rather than re-derive it. Consequently,
\begin{equation}
L_{\text{reuse}}(N) = O\!\left(\frac{\log N}{\log \log N}\right).
\end{equation}

\subsection{A Finite-$N$ Refinement}

The asymptotic form of Theorem~\ref{thm:maxdegree} converges slowly:
at the swarm sizes tested in Section~\ref{sec:results}, it
underestimates simulated values by a roughly constant multiplicative
factor. A tighter finite-$N$ estimate solves directly for the
smallest $k$ such that $N \cdot \Pr[\text{Poisson}(\mu) \ge k] \le 1$,
i.e., inverting the Poisson survival function rather than using the
leading-order asymptotic term alone. We report both the asymptotic
bound (the formal, citable claim) and this refined estimate (which
matches simulated results closely) in Section~\ref{sec:results}, and
recommend the latter for any reader wishing to sanity-check finite-$N$
simulation output against theory.

\subsection{Static Baseline}

We compare against 2-D carrier-hopping prime codes (CHPC)
\cite{kwong2020automotive}, the established static baseline in
automotive/robotic time-of-flight LiDAR OCDMA. This baseline is
chosen not because it yields a different asymptotic growth rate than
simpler code families (Gold codes are also $\Theta(N)$), but because
it is the physically validated, field-standard baseline directly from
the closest prior art, and because it decouples code weight from
code length more efficiently than one-dimensional alternatives,
making it the strongest legitimate comparison rather than a weak
strawman. Critically, CHPC cardinality equals chip length exactly --
not $\Theta(\sqrt{N})$ or any other sublinear relationship -- verified
directly from the cited source: a CHPC of chip length $N{=}31$
supports exactly 31 sequences. Hence
\begin{equation}
L_{\text{static}}^{\text{CHPC}}(N) = N \quad \text{(exact, not asymptotic).}
\end{equation}

\begin{corollary}
\label{cor:gap}
\begin{equation}
\frac{L_{\text{static}}^{\text{CHPC}}(N)}{L_{\text{reuse}}(N)} = \Theta\!\left(\frac{N \log\log N}{\log N}\right) \to \infty \quad \text{as } N \to \infty.
\end{equation}
\end{corollary}
\noindent The advantage of dynamic reuse over static assignment is
therefore unbounded, not merely a constant-factor improvement.

\subsection{Scope of the Proof}

Theorem~\ref{thm:maxdegree} and Corollary~\ref{cor:gap} bound code
demand under the true, instantaneous interference graph $G(t)$ --
that is, under perfect, immediate neighbor knowledge. The protocol
actually operates on the estimated graph derived from
$\hat{\mathcal{N}}_i(t)$, which lags and can err (beacon period
$T_b$, missed-neighbor rate $p_{fn}$, phantom-neighbor rate $p_{fp}$).
$L_{\text{reuse}}(N)$ as derived here is therefore a lower bound on
the code demand actually required in practice; the gap between this
bound and operational reality is quantified empirically in
Sections~\ref{sec:collision}--\ref{sec:results}, and is not claimed to
be closed by this proof alone.

\section{Collision Model}
\label{sec:collision}

Section~\ref{sec:scaling} bounds code demand under idealized
detection. We now characterize how imperfect, beacon-limited
detection produces the collision events defined in
Section~\ref{sec:sysmodel} in practice -- decomposing the gap between
a conflict and the collision it fails to prevent into three distinct
mechanisms.

\subsection{Three Collision Types}

\begin{itemize}
\item \textbf{Type I (silent miss):} a true neighbor $j \in
\mathcal{N}_i(t)$ is absent from $\hat{\mathcal{N}}_i(t)$ due to
beacon loss, probability $p_{fn}$ per beacon.
\item \textbf{Type II (detection lag):} robot $j$ enters range $D$ of
robot $i$ at time $t_0$; even with zero detection error, $j \notin
\hat{\mathcal{N}}_i(t)$ until the next successful beacon at
$t_0 + \tau_{det}$. This is structural: it persists even as
$p_{fn}, p_{fp} \to 0$, so long as $T_b > 0$.
\item \textbf{Type III (handoff-transient):} a robot reassigns to
resolve a detected conflict, and the newly selected code coincides
with a third, previously unrelated robot outside its current
detection horizon.
\end{itemize}

\subsection{Encounter Duration}

For two robots sharing a common aisle in the warehouse mobility model
of Section~\ref{sec:sim-methodology}, relative motion is
one-dimensional, and the duration of a close encounter admits a
simple closed form from vehicular-network connectivity-duration
analysis \cite{vanet-velocity}:
\begin{equation}
t_c = \frac{2D}{|\Delta v|},
\end{equation}
where $\Delta v$ is the relative speed between the two robots. This
governs the exposure window relevant to Type II.

\subsection{Grounded Parameters}

Table~\ref{tab:params} reports the parameter values used throughout
Sections~\ref{sec:sim-methodology}--\ref{sec:results}, each anchored
to a measured source rather than an assumed or invented value, with
the detection-error rate treated as a sensitivity sweep rather than a
single point estimate given its documented dependence on deployment
geometry.

\begin{table}[t]
\centering
\caption{Grounded simulation parameters}
\label{tab:params}
\begin{tabular}{@{}lll@{}}
\toprule
Parameter & Value / range & Source \\
\midrule
Robot speed $\bar{v}$ & 1.5 m/s (top speed) & AGV industry spec \\
Beacon period $T_b$ & $\ge$ 100 ms & UWB ranging hardware \\
Position noise $\sigma_{pos}$ & 0.025--3.78 m & UWB LOS/NLOS studies \\
$p_{fn}, p_{fp}$ & swept 0--0.25 & anchored NLOS incidence \\
\bottomrule
\end{tabular}
\end{table}

\subsection{Collision Probability}

The per-robot, per-unit-time collision probability satisfies the
union bound
\begin{equation}
P_{\text{collision}}(t) \le P_I(t) + P_{II}(t) + P_{III}(t),
\end{equation}
where $P_I$ and $P_{II}$ are approximated analytically from the
neighbor-arrival rate and $p_{fn}$ respectively, and $P_{III}$ is
characterized empirically in Section~\ref{sec:results} rather than
bounded in closed form -- the handoff-transient case depends on
third-party state outside any single robot's detection horizon, and
we report this as an explicit, deliberate scope decision rather than
an omission.

\section{Simulation Methodology}
\label{sec:sim-methodology}

To evaluate the protocol under realistic conditions, we implement a
discrete-time simulator with four validated components: an
interference-graph and code-assignment layer, a graph-constrained
mobility layer, an imperfect-detection layer, and a Monte Carlo
harness. Each component was individually validated against known
ground truth before composition, described below along with two
methodological corrections made during development, reported here in
the interest of reproducibility.

\subsection{Warehouse Layout}

\begin{figure}[t]
\centering
\includegraphics[width=0.9\linewidth]{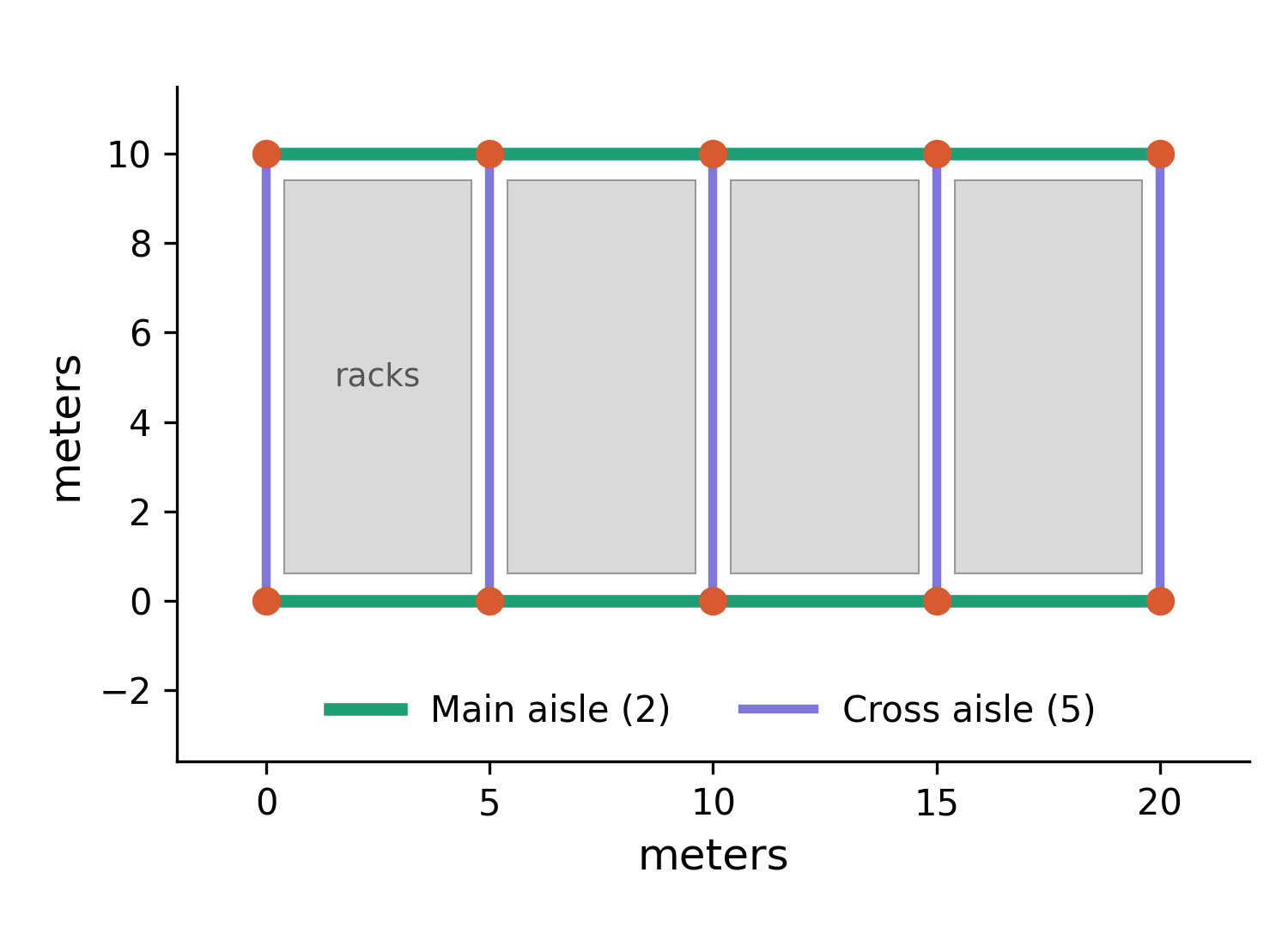}
\caption{Warehouse aisle layout: two main aisles, five cross aisles, ten graph nodes. Racking (gray) blocks direct motion between aisles.}
\label{fig:warehouse}
\end{figure}

The environment is a $20\,\text{m} \times 10\,\text{m}$ rectangular
tile: two main aisles along the long edges (10 m apart) and five
cross aisles connecting them at 5 m intervals, forming a $2\times5$
grid of ten intersections (Fig.~\ref{fig:warehouse}). Racking between
aisles physically blocks direct motion, which is the justification
for constraining robot mobility to this graph rather than open
two-dimensional motion. For Regime A (Section~\ref{sec:scaling}), the
tile is repeated along its length, sharing boundary aisles between
adjacent tiles, to grow operating area while holding local aisle
density fixed.

\subsection{Mobility Model}

Robots move along aisle edges at fixed speed $\bar{v}$, selecting a
random adjacent edge at each intersection -- a graph-constrained
analogue of random-waypoint motion, chosen over open-plane random
waypoint because the latter is known to produce non-uniform spatial
density (concentration toward the region center), which would
conflict with the constant-density assumption of
Section~\ref{sec:scaling}.

\subsection{Detection Error Model}

Each robot's belief about its neighbors updates only at beacon
intervals $T_b$ and is corrupted by $p_{fn}$ (a true neighbor is
dropped from the update) and $p_{fp}$ (a phantom neighbor is added).
During development, the simulation step size $\Delta t$ was initially
set coarser than $T_b$, which caused the beacon to fire on every
simulation step regardless of the configured period, silently
erasing detection lag from the model; this was caught by observing
that collision rate was statistically insensitive to $p_{fn}$ across
its entire tested range, corrected by enforcing $\Delta t < T_b$, and
confirmed by re-testing the sweep.

\subsection{Monte Carlo Discipline}

All reported results use $\ge 30$ independent random seeds per
condition, with 95\% confidence intervals computed via the normal
approximation. Step size $\Delta t$ was separately validated by
convergence testing for every metric reported: collision counting was
initially implemented as a per-check event tally, which does not
converge as $\Delta t \to 0$ (a fixed-duration collision is counted
once per check, so the tally scales as $1/\Delta t$); this was
corrected by tracking collision \emph{episodes} (a transition into a
colliding state) and total collision \emph{time} (which converges to
the true duration as $\Delta t$ shrinks) instead of a raw count. Peak
simultaneous code usage, by contrast, is a running maximum and
converges properly under refinement without correction, which was
confirmed empirically before being relied upon as a metric.

\subsection{Baseline Construction}

We compare against a simplified, structurally faithful model of an
existing decentralized, coordination-free interference-mitigation
approach for LiDAR \cite{rathnayake2024d2sr}, rather than reproducing
its full sensor-level detection pipeline, which is out of scope. The
structural property that matters for comparison is the presence or
absence of coordination: the baseline has no beacon and no neighbor
list, only a per-robot self-diagnosis of whether it is currently
experiencing interference, with detection reliability grounded in the
source's own reported recall ($\approx 82$--$83\%$) and precision
($\approx 99\%$, implying a $\approx 1\%$ false-alarm rate). On
detecting interference, the baseline reassigns \emph{blindly} -- a
uniform random draw with no informed exclusion -- in contrast to our
protocol's informed, exclusion-based selection. Both protocols share
the identical warehouse layout, mobility model, and physical
interference check, isolating the comparison to this one structural
difference. An initial version of this comparison omitted the
baseline's false-alarm rate and allowed the two protocols an unequal
effective code-search window, both of which favored the baseline
artificially; both were corrected prior to the results reported in
Section~\ref{sec:results}.

\section{Results}
\label{sec:results}

\subsection{Theory-Simulation Agreement}

\begin{figure}[t]
\centering
\includegraphics[width=0.9\linewidth]{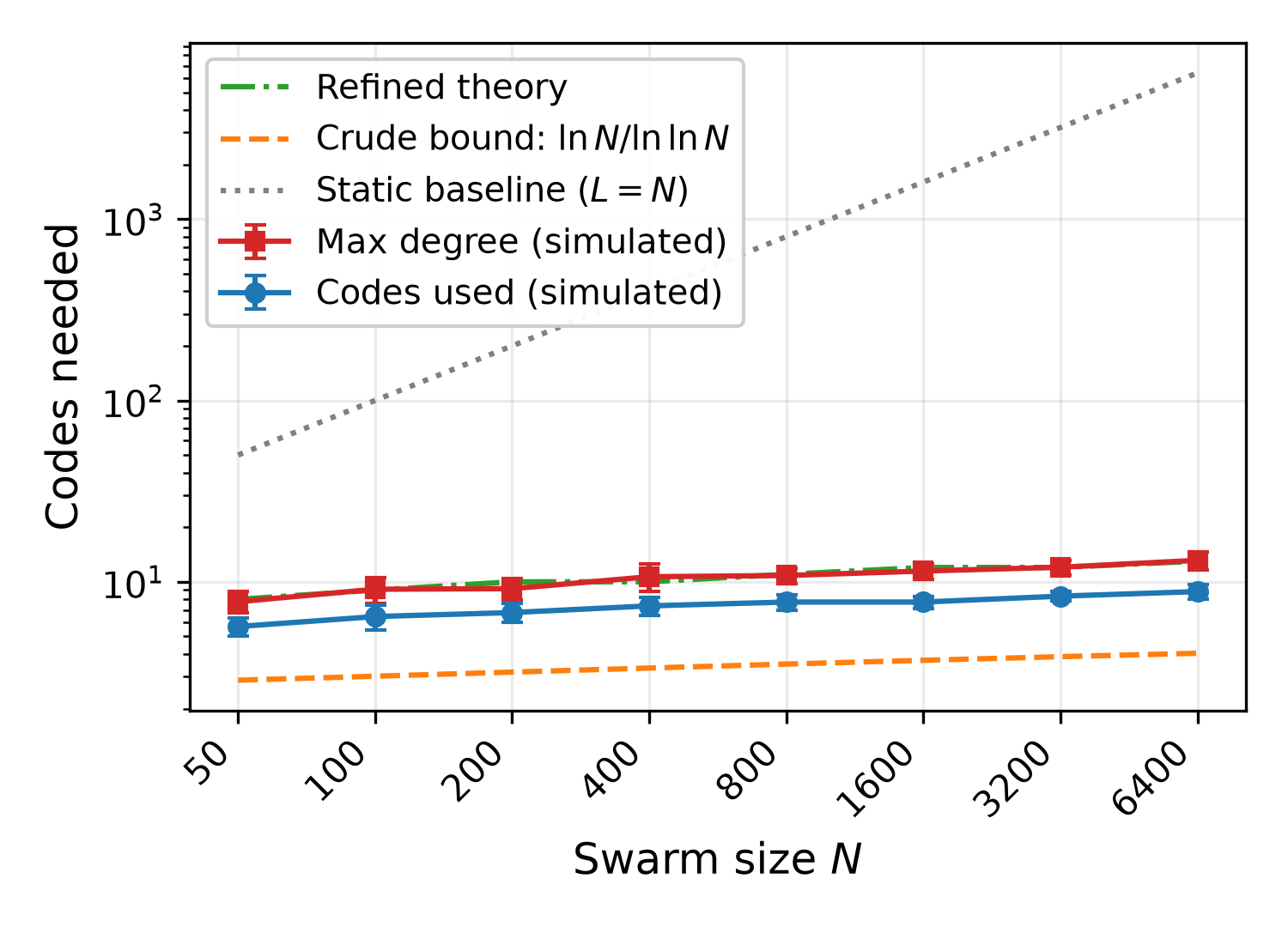}
\caption{Codes required: simulated (mean $\pm$ 1 std, 30 seeds), theoretical bounds, and static baseline, $N \in [50, 6400]$.}
\label{fig:piece1}
\end{figure}

Fig.~\ref{fig:piece1} confirms Theorem~\ref{thm:maxdegree} and its
finite-$N$ refinement against simulated maximum degree across
$N \in [50, 6400]$, 30 seeds per point. The refined estimate matches
simulated values closely; the leading-order asymptotic alone
underestimates by a roughly constant multiplicative factor across
this range, consistent with the known slow convergence of
$\log/\log\log$-type bounds.

\subsection{Realistic Headline Result}

\begin{figure}[t]
\centering
\includegraphics[width=0.9\linewidth]{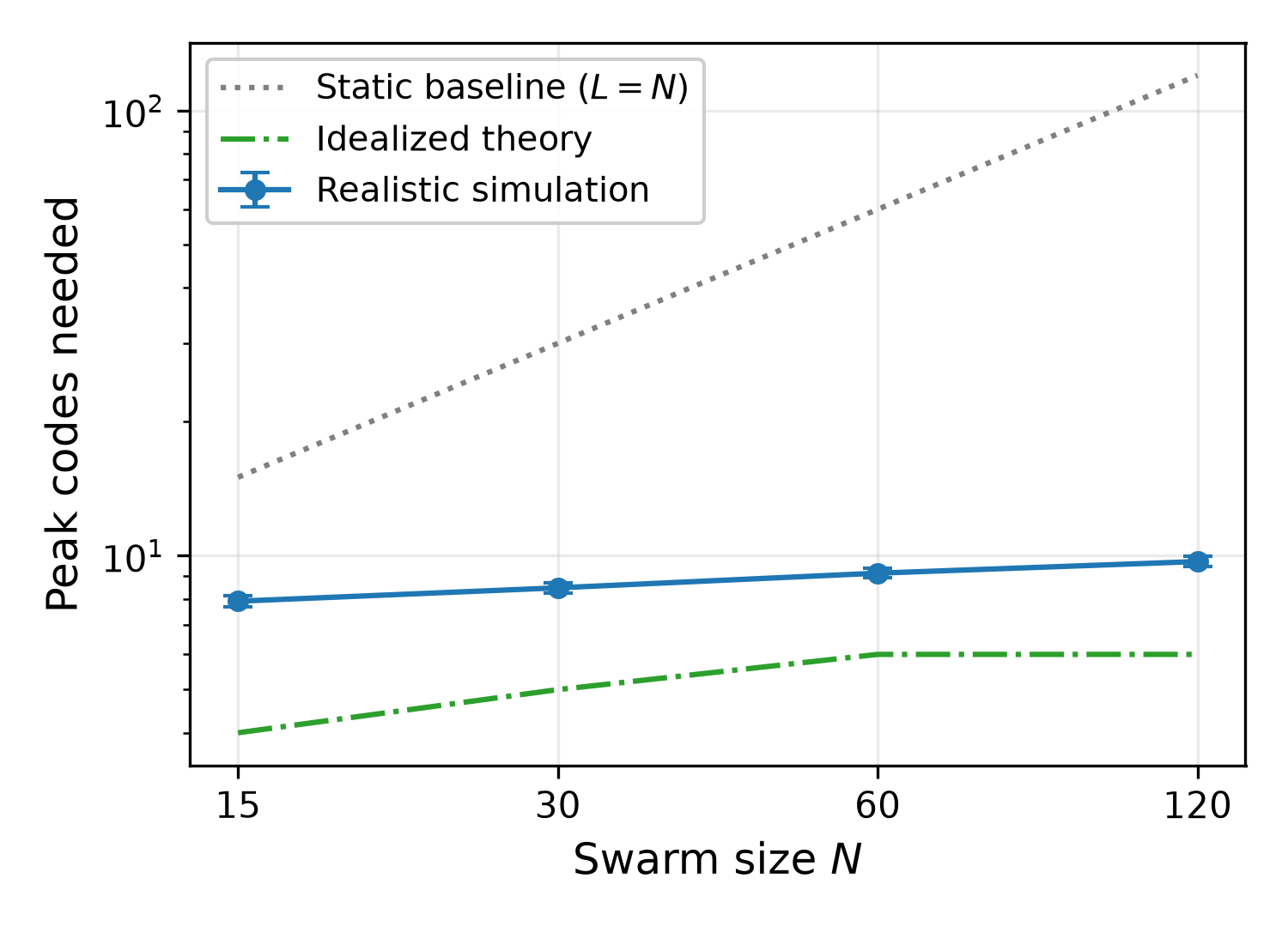}
\caption{Peak codes needed under full realism vs. swarm size, against the idealized theoretical reference and the static baseline.}
\label{fig:piece5}
\end{figure}

Fig.~\ref{fig:piece5} reports peak simultaneous code usage under full
realism -- mobility, worst-case detection error ($p_{fn}{=}0.25$),
and reactive reassignment -- across $N \in \{15,30,60,120\}$
(1, 2, 4, 8 warehouse tiles), 30 seeds per point. The gap between our
protocol and the static baseline widens with $N$: a 2$\times$
reduction at $N{=}15$ grows to a 12$\times$ reduction at $N{=}120$,
consistent with the widening, unbounded gap of Corollary~\ref{cor:gap}
under realistic conditions rather than only under the idealized
theory of Section~\ref{sec:scaling}.

\subsection{Comparison Against a Coordination-Free Baseline}

\begin{figure}[t]
\centering
\includegraphics[width=0.9\linewidth]{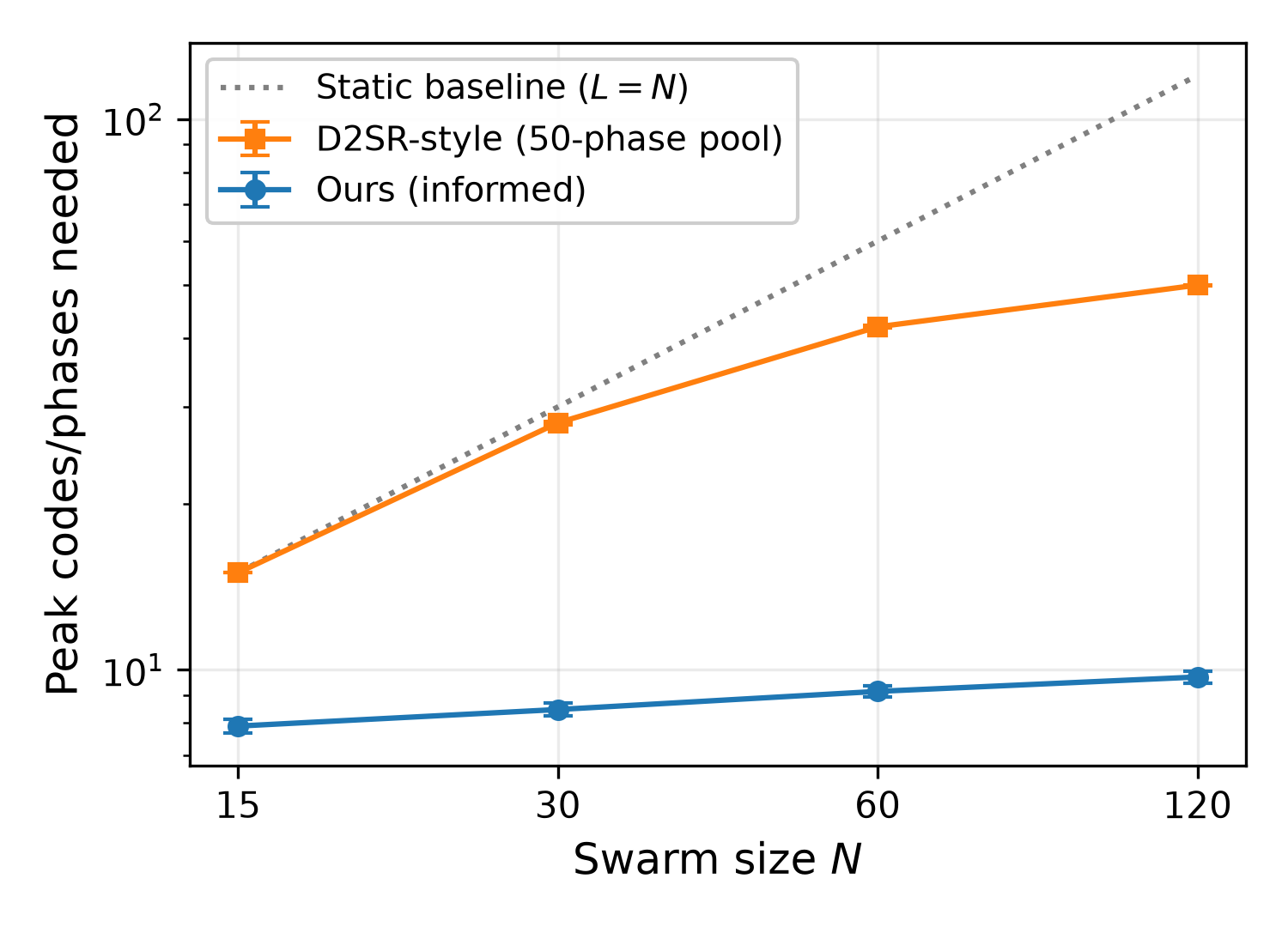}
\caption{Resource efficiency: peak codes/phases needed, generous shared pool.}
\label{fig:d2sr-eff}
\end{figure}

\begin{figure}[t]
\centering
\includegraphics[width=0.9\linewidth]{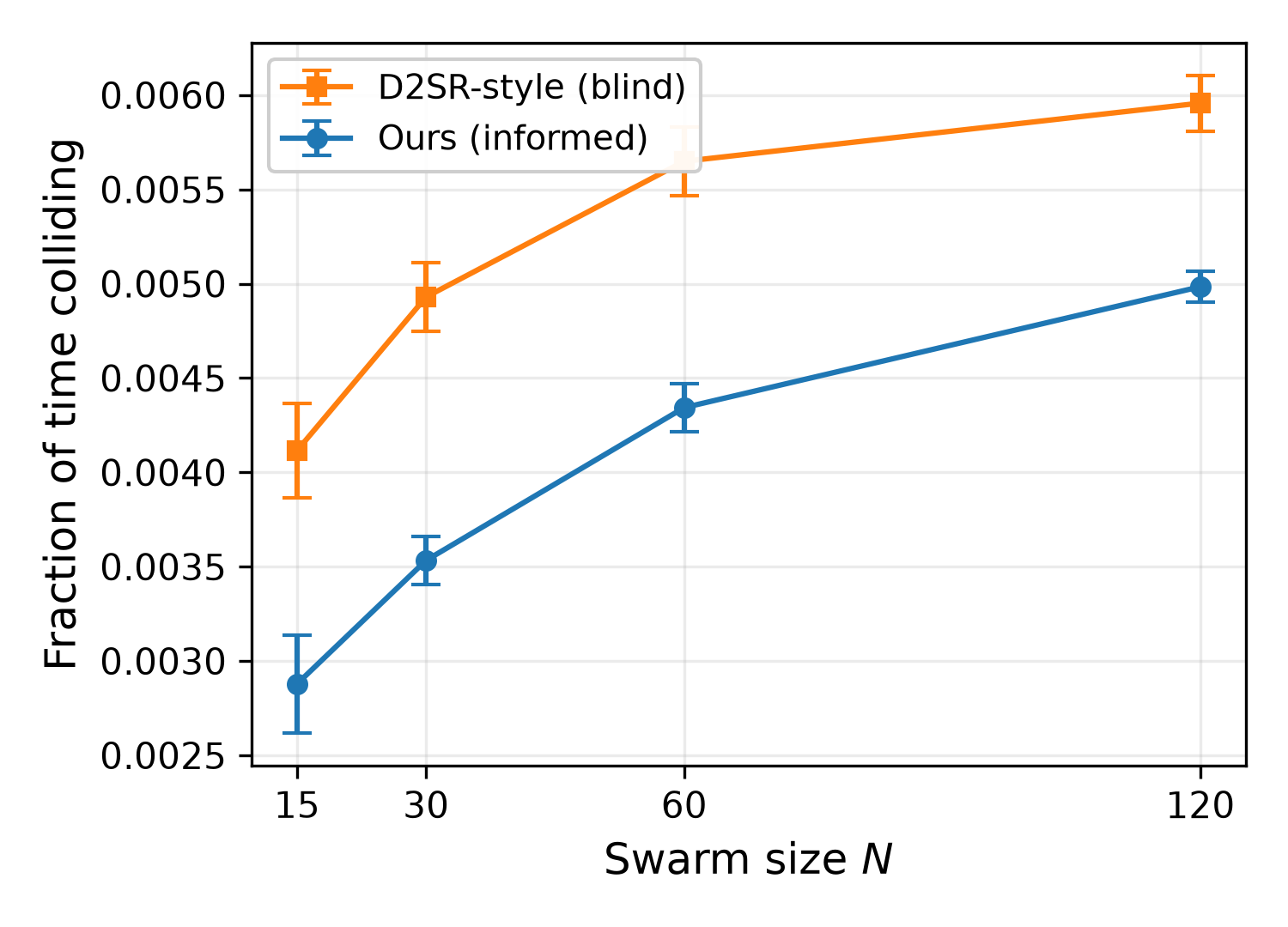}
\caption{Collision-time fraction under an identical, constrained code budget.}
\label{fig:d2sr-equal}
\end{figure}

Fig.~\ref{fig:d2sr-eff} shows that the coordination-free baseline
achieves little genuine reuse under a generous shared code pool,
tracking close to the static baseline rather than our protocol.
Fig.~\ref{fig:d2sr-equal} shows that under an identical, constrained
code budget matched to what our protocol requires, informed,
coordinated selection yields consistently lower collision-time
fraction than blind selection across the same swarm-size sweep.

\section{Discussion and Limitations}
\label{sec:discussion}

We identify five scope limitations of this work, each defining a
concrete direction for refinement rather than an open question left
unaddressed.

\textbf{Interference checking assumes line-of-sight.}
Sections~\ref{sec:sim-methodology}--\ref{sec:results} determine
interference from Euclidean separation $d_{ij}(t) \le D$ alone,
without modeling whether racking between aisles would occlude the
direct path between two robots. Given the 5\,m cross-aisle spacing
used throughout, robots in adjacent aisles remain plausibly within
physical interference range regardless of intervening structure, but
a full geometric line-of-sight model -- resolving occlusion directly
from rack placement in the warehouse graph -- would tighten this
assumption and is a natural simulator extension.

\textbf{Only Regime A is analyzed.} Section~\ref{sec:scaling} proves
the scaling result under constant robot density with growing
operating area -- an expanding or exploring swarm. The complementary
regime, fixed operating area with growing density, characteristic of
an increasingly crowded fixed footprint, requires a distinct proof,
since local vertex degree is no longer independent of $N$, and is
left to future work.

\textbf{One concrete warehouse layout.} The aisle graph of
Section~\ref{sec:sim-methodology} is a single, representative
instance of a standard main-aisle/cross-aisle design, not an
exhaustive exploration of layout sensitivity. Alternative topologies
-- irregular aisle spacing, non-rectangular footprints, multiple
main-aisle pairs -- may alter the neighbor-arrival-rate constant left
unresolved in Section~\ref{sec:collision}, though the scaling result
of Section~\ref{sec:scaling} does not depend on this specific layout.

\textbf{The baseline is a simplified, structurally faithful model,
not a full reproduction.} The coordination-free baseline of
Section~\ref{sec:sim-methodology} isolates the presence or absence of
inter-robot coordination -- the structural property relevant to this
paper's claim -- using the source's own reported detection
reliability, rather than reimplementing its underlying sensor-level
detection pipeline, which operates at a different level of
abstraction and is outside this paper's scope.

\textbf{Validation is simulation-only.} All results in this paper are
obtained via Monte Carlo simulation; no physical hardware validation
is presented. This is standard for early-stage protocol contributions
at this venue, but physical deployment -- subject to real beacon
hardware, uncontrolled RF interference, and sensor noise beyond what
is modeled in Table~\ref{tab:params} -- remains necessary before an
operational deployment claim could be made.

These limitations, together with the synchronous-beaconing assumption
of Section~\ref{sec:protocol}, motivate the future directions
outlined in Section~\ref{sec:conclusion}.

\section{Conclusion}
\label{sec:conclusion}

We presented a decentralized protocol for dynamic, spatial code reuse
in LiDAR-equipped robot swarms, replacing static, per-robot code
assignment with reassignment driven by a live, beacon-maintained
interference-neighborhood graph. We proved that the number of codes
required grows as $O(\log N/\log\log N)$ under constant robot
density -- an unbounded improvement over the $\Theta(N)$ growth
required by static assignment -- and validated this result under
conditions substantially more adversarial than the idealized proof:
robot mobility, imperfect beacon-based detection, and reactive
reassignment. Against a structurally faithful, fairly-constructed
model of an existing coordination-free approach, our protocol
achieved both greater code efficiency and lower collision risk under
an identical resource budget, directly demonstrating that
coordination -- not merely reactivity -- is what closes the scaling
gap.

The limitations identified in Section~\ref{sec:discussion} motivate
several directions for future work:

\begin{itemize}
\item \textbf{Dense-swarm scaling (Regime B).} A fixed operating
footprint with growing robot density requires a distinct proof from
the one presented here, and is planned as a direct follow-on to this
work.
\item \textbf{Asynchronous beaconing.} Relaxing the synchronous,
clock-aligned beacon assumption of Section~\ref{sec:protocol} toward
event-driven coordination, more representative of a deployed swarm
without hardware clock synchronization.
\item \textbf{Refined environmental modeling.} Full geometric
line-of-sight occlusion and a broader sweep of warehouse layout
topologies, both identified in Section~\ref{sec:discussion}.
\item \textbf{Hardware validation.} Deployment on physical robots,
subject to real beacon hardware and uncontrolled RF conditions, as
the natural next step toward an operational deployment claim.
\item \textbf{Longer-horizon extensions.} Adversarial, rather than
average-case, detection failure; three-dimensional aerial swarms,
which alter the underlying neighbor-graph degree scaling; and a
learned, rather than fixed, reuse policy.
\end{itemize}

\section*{Acknowledgment}

The author used Claude (Anthropic) as an AI assistant in the
preparation of this manuscript, including drafting of the manuscript
text across all sections and implementation of the Python simulation
code described in Section~\ref{sec:sim-methodology}. All research
design decisions, experimental configurations, execution of
simulations, interpretation of results, and verification of cited
sources against primary references were performed by the author. The
author has reviewed and takes full responsibility for the accuracy,
originality, and integrity of all content in this paper.

\bibliographystyle{IEEEtran}

\end{document}